\documentclass[sigconf]{acmart}
\usepackage[inkscapeformat=png]{svg}
\graphicspath{{figures/}}
\usepackage{graphicx}
\usepackage{tabularx}
\usepackage{multirow}
\usepackage{marvosym}
\usepackage{pdfpages}
\usepackage{subfiles}
\usepackage{colortbl}

\AtBeginDocument{%
  \providecommand\BibTeX{{%
    \normalfont B\kern-0.5em{\scshape i\kern-0.25em b}\kern-0.8em\TeX}}}

\setcopyright{acmcopyright}
\copyrightyear{2023}
\acmYear{2023}
\acmDOI{XXXXXXX.XXXXXXX}

\acmConference[CBMI '23]{}{Orleans}{France}
\begin{document}

    \title[Detecting OOC Image Caption Pair in News: A Counter-Intuitive Method]{Detecting Out-of-Context Image-Caption Pairs in News:\\A Counter-Intuitive Method}

\author{Eivind Moholdt}
\affiliation{%
  \institution{University of Bergen}
  \city{Bergen}
  \country{Norway}}
\email{buh011@uib.no}

\author{Sohail Ahmed Khan}
\affiliation{%
  \institution{University of Bergen}
  \city{Bergen}
  \country{Norway}}
\email{sohail.khan@uib.no}

\author{Duc-Tien Dang-Nguyen}
\affiliation{%
  \institution{University of Bergen}
  \city{Bergen}
  \country{Norway}}
\email{ductien.dangnguyen@uib.no}

\renewcommand{\shortauthors}{Moholdt et al.}

\begin{abstract}
The growth of misinformation and re-contextualized media in social media and news leads to an increasing need for fact-checking methods. Concurrently, the advancement in generative models makes cheapfakes and deepfakes both easier to make and harder to detect. In this paper, we present a novel approach using generative image models to our advantage for detecting Out-of-Context (OOC) use of images-caption pairs in news. We present two new datasets with a total of $6800$ images generated using two different generative models including (1) DALL-E 2, and (2) Stable-Diffusion. We are confident that the method proposed in this paper can further research on generative models in the field of cheapfake detection, and that the resulting datasets can be used to train and evaluate new models aimed at detecting cheapfakes. We run a preliminary qualitative and quantitative analysis to evaluate the performance of each image generation model for this task, and evaluate a handful of methods for computing image similarity.  
\end{abstract}


\begin{teaserfigure}
    \centering
    \includegraphics[width=.9\linewidth]{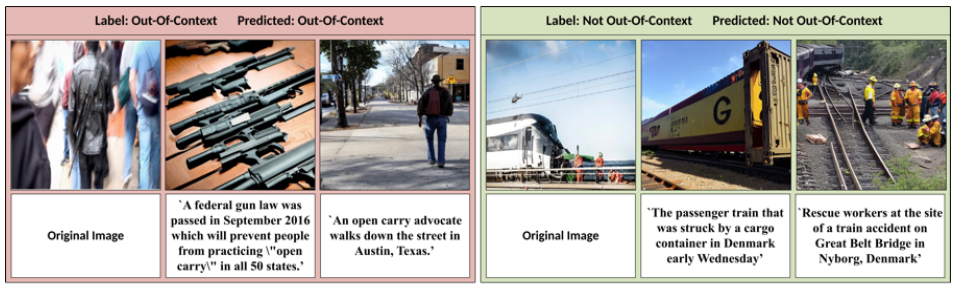}
  \caption{The core idea of this paper is to utilize the perceptual similarity between synthetically generated images (from both DALLE-2 and Stable Diffusion) for detecting Out-Of-Context content in image and caption pairs.}
  \Description{Photo grid of original image compared to AI generated images}
  \label{fig:teaser}
\end{teaserfigure}

\maketitle

\section{Introduction}
While the increase in volume and accessibility of information available online has the potential to improve education, communication, and decision-making, it has also created a fertile breeding ground for misinformation, fake news, and propaganda. Concurrently, the recent advancement in generative models enables a new method of spreading misinformation via AI generated images and videos. The accelerating growth of false and misleading information online is a major threat which could lead to increased confusion and distrust in media or political institutions \cite{cheapfake_schick}. Moreover, with the advent of deep-learning technologies such as 'deepfakes'  or generative models that can produce near realistic images from text prompts, deepfakes are getting increasingly sophisticated and harder to detect, making it more difficult for individuals to distinguish between true and false information \cite{goldstein2023generative}.

\begin{figure*}[t!]
    \centering
    \includegraphics[width=0.9\linewidth]{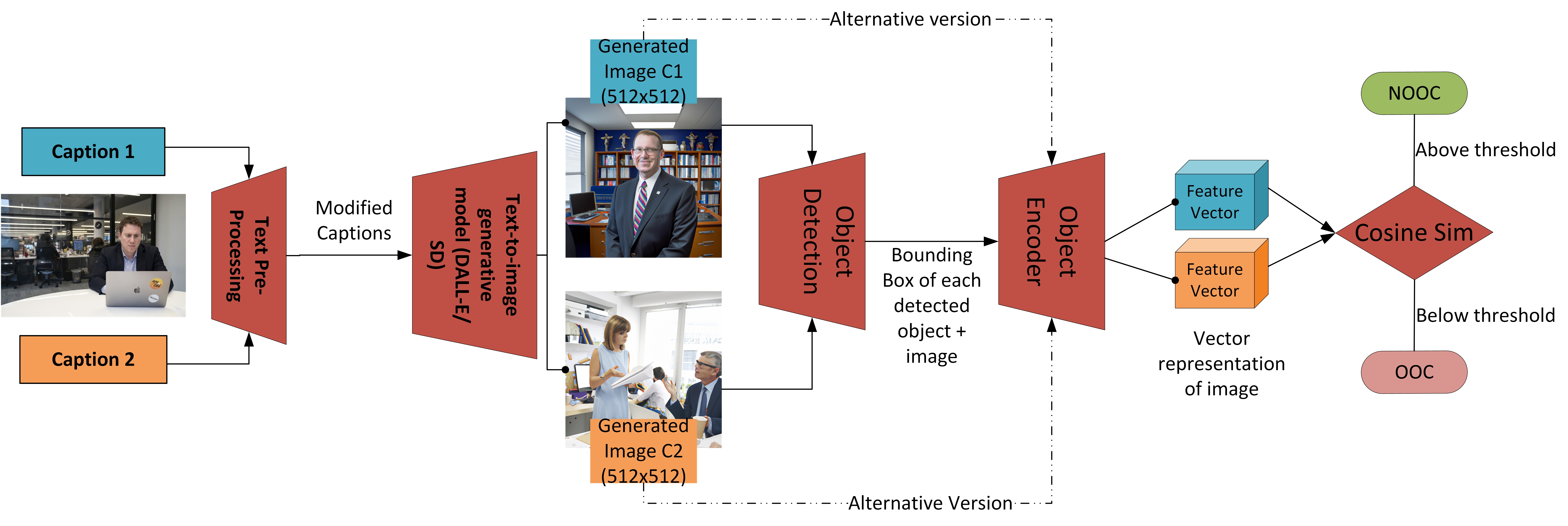}
    \caption{High-level architecture of the proposed model}
    \label{fig:model}
\end{figure*}

Cheapfake media refers to manipulated or fake media created using `cheap' tools such as photo or video editors to alter images, videos, or text. This type of misinformation is easy to create and can be spread quickly online, and is also difficult to detect. Media that includes images are more likely to be clicked and shared, and statements are more often believed when presented alongside an image, making cheapfakes a particular effective way to spread misinformation online \cite{fazioOOC}. Preceding work of Aneja \textit{et al.} \cite{COSMOS} in the COSMOS project presented the first automatic cheapfake detection model in order to detect mismatch in image and captions, along with a large-scale dataset of 200K images with 450K textual captions. The test set includes image-caption triplets that are labeled Out-Of-Context (OOC) or Not-Out-Of-Context (NOOC) \cite{COSMOS}. 

Given the recent advancements in text-to-image generative models, we propose that they can be employed for cheapfake detection by generating images that express the caption's content. In this paper, we present two new datasets comprising of 3400 images each, generated using OpenAI's DALL-E 2 \cite{dalle2}\cite{ramesh2022hierarchical}, and Stable Diffusion \cite{compvis/stable-diffusion-v1-4huggingface}\cite{Rombach_2022_CVPR} along with the textual captions used to generate the images.  We also present a novel approach for verifying the consistency between captions and images by comparing the perceptual similarity between AI generated images based on the captions from the COSMOS dataset. The idea behind our approach is that, synthetically generated images (from both DALL-E 2 and Stable Diffusion) should have a high semantic similarity towards each other or the original image if the captions are also similar in their semantics, and thus can help identify OOC cheapfake media. We carry out qualitative as well as quantitative analysis of our proposed method on the generated datasets, and report some insightful results in our study. The code and the generated datasets are available at the \href{https://github.com/eivindmoholdt/Master-Code.git}{Github Repository} \footnote{https://github.com/eivindmoholdt/Master-Code.git}.

\section{Related Work}
New cheapfake detection models have improved the accuracy of the COSMOS baseline in several different ways. Akgul \textit{et al.} propose a method named Differential Sensing, which adds a negative (‘was not true’) and positive (‘was true') probe to each caption \cite{akgul2021cosmos}. Using the SBERT similarity score between the original captions and the probes, the scores moving opposite directions when compared to the original captions would indicate the captions contradicted each other \cite{SBERT}. This method increased the accuracy from 81.9\% to 85.6\% on the test set. Tran \textit{et al.} propose a Natural Language Inference (NLI) task to determine whether a given caption pair contradict or entail each other in order to address the relationship between the captions. 
They also propose a Online Caption Checking method that crawls online resources to find a third caption to gain additional context to an image and verify the caption’s truthfulness. This method however struggled to verify the captions, as the third caption sometimes would relate to a different image in the article with the original image  \cite{tran2022textual}. La \textit{et al.} demonstrates a bottom-up attention model with visual semantic reasoning to extract image features for image-text matching, resulting in a 4.5\% increase from the COSMOS baseline \cite{tuan2022combi}. 

\subsection{Stable Diffusion}
Stable Diffusion is a latent text-to-image diffusion model developed by Stability AI and published in 2022 \cite{Rombach_2022_CVPR}. It combines deep learning techniques and probabilistic modeling to generate high-quality images. Unlike basic diffusion models, Stable Diffusion operates in the latent space, where it applies noise to a compressed representation of the data and then performs denoising operations to recover samples in the data space. This approach is computationally efficient while preserving the essential features of the image. The model consists of three main components: a Variational AutoEncoder (VAE) encoder, a U-Net block with a ResNet backbone for denoising, and a VAE decoder that generates the final image from the reconstructed latent representation \cite{Rombach_2022_CVPR}.

\subsection{DALL-E 2}
DALL-E 2 is a text-to-image model developed by OpenAI \cite{ramesh2022hierarchical}. It combines two previously published models: CLIP and GLIDE. CLIP is a zero-shot neural network introduced in 2021, which forms the foundation for the multimodal approach to image synthesis. It learns to associate objects in sentences with objects in images using contrastive learning, creating a joint embedding space for visual and textual information. CLIP is also robust to distribution shifts, making it generalize well to different data patterns \cite{radford2021learning}. GLIDE, proposed in 2022, modifies the basic diffusion model to incorporate CLIP during training. By augmenting the training process with CLIP text embeddings, GLIDE enables text-conditional image generation, producing images that align better with the text's semantics \cite{nichol2022glide}. DALL-E 2 consists of three main components: a text encoder (CLIP) that maps text to an embedding, a diffusion model (prior) that maps the text encoding to an image encoding, and the GLIDE model that decodes the image encoding into the final image. This process is referred to as unCLIP by OpenAI \cite{ramesh2022hierarchical}.

\section{Proposed Method}
We propose a novel approach for detecting OOC captions and image pairs by comparing the perceptual similarity between AI generated images based on the captions from the COSMOS dataset. As image generation is time-consuming, we present two new datasets of synthetically generated images along with their textual captions that can be used for future research in this area. We employ two newly proposed synthetic text-to-image generative models, (1) Stable Diffusion~\cite{Rombach_2022_CVPR, compvis/stable-diffusion-v1-4huggingface} and (2) DALL-E 2 \cite{ramesh2022hierarchical, dalle2} to generate the datasets. Both models report state-of-the-art performance on benchmarks for image generation models, and are able to both generate highly realistic images, as well as images that have a high semantic alignment towards the input caption. Thus, both models are suitable for this task. We perform a qualitative analysis of the generated images by conducting a user study to achieve annotated similarity ratings, as well as a quantitative analysis to test the effectiveness of the proposed method. We employ a feature-based approach for computing image similarity, utilizing an object detection model and an object encoder to capture the semantic content of the images into feature vector representations, and computing their similarity with Cosine Similarity. By computing the similarity of the original image vs each generated image, or the similarity of the generated images, we predict OOC/NOOC and verify our results against the gold labels from the COSMOS dataset \cite{COSMOS}.

\subsection{Pre-processing}
Before prompting the captions to the image generation models, extensive text pre-processing is needed. The COSMOS captions often include political statements, slurs, fake news and misleading information. This provides a challenge in order to comply with the ethical use and content policy filters of the image generation models. OpenAI's content policy for DALL-E 2 states that the user is not allow to `create, upload, or share images that are not G-rated or that could cause harm' \cite{dalle2}. This includes names, foul language, violence, drugs, and more. In addition, DALL-E 2's safety filter is also activated by topics such as COVID-19, abortion, pregnancy, drugs and more, which are necessary to remove from the dataset. The former provides a challenge as seven \% of the COSMOS dataset falls under the `Covid' category \cite{COSMOS}. Extensive text-processing is therefore needed before using the prompts to generate images. We use the modified captions from the COSMOS dataset that has been pre-processed using Named Entity Recognition (NER) to replace proper nouns with corresponding entity labels, such as replacing `Obama' with `Person'. NER also helps decrease the abstraction between the caption and the images, as neither of the image generation models or object detection models will distinguish between types of persons or locations. We believe this will make similarity comparisons easier, thus increasing the accuracy of our model. An additional round of NER processing is performed to identify any proper nouns that were missed during the initial step using the same en\_core\_web\_sm model from the spaCy library as COSMOS. Furthermore, a list of inappropriate words is compiled and used as an additional filter before prompting the captions to DALL-E 2 and Stable Diffusion. As a base for our list, we utilize the open-source Github LDNOOBW list from Shutterstock \cite{LDNOOBW}.

\subsection{Image generation and dataset collection}
For this project we use the Test set from the COSMOS dataset, which includes 1700 images and 3400 captions. We generate one synthetic image for each caption corresponding to the original image, gaining 3400 generated images in each dataset, totaling 6800 generated images. We generate images of 512x512 pixels for each model to ensure comparability between the datasets. This can be easily changed for both models. Generating images with higher resolution will be more expensive and time-consuming. Due to the generative nature of the models, providing the same prompt to the model twice will results in a different generated image. Thus, even captions that are completely similar will produce different results. While this can produce variations in the output, we still anticipate a high degree of similarity in captions that possess semantic similarity.  Although the process is automated using Python, generating images is time-consuming: Stable Diffusion model uses about 15 seconds to generate each image and save it to our directory with standard Google Colab GPU. DALL-E 2 uses about seven seconds to generate each image and save it to our directory with a standard Google Colab GPU. To speed up the process, we utilize premium GPU from Colab for the Stable Diffusion model, which decreased the generation runtime to three seconds per image. For this reason, verifying our proposed method might be hard if images are to be generated in real time. In order to facilitate for further testing we therefore present the datasets in this paper.

\subsection{Computing Image Similarity}
There are several algorithms available for computing image similarity, such as pixel-to-pixel comparison with MSE or structural comparison using SSIM. Given that image generation models introduce randomness, resulting in differences in object placement and angles in the generated images, traditional pixel-to-pixel or SSIM comparison may not be effective. To capture the context of the images, we use a feature-based similarity approach: Our method uses feature extraction techniques to extract high-level features from images to vector representations, which are then used to compute the similarity between images with Cosine Similarity. Using object encoders such as ResNet50, we can extract features from images and compare them using distance metrics such as Cosine similarity. We test 8 different object encoders for this task. We also employ 3 object detection models and combine them with all object encoders to find the best method for capturing the semantic content of the images for image similarity comparison. 




Our prediction model employs two methods. In the first method, we use a pre-trained object detection model to locate objects in an image and create bounding boxes around them. The idea is that using a combination of an object detection model and an object encoder, we can capture contextual information and relationships between objects, resulting in a more accurate feature vector representation of the image. While object encoders also includes object detection capabilities, object detection models such as YOLO and MASK-RCNN are specialized for this, and have a higher detection accuracy and more accurate classification capabilities. The bounding boxes are used to isolate each object as a separate image, which is then processed by the object encoder. The feature vectors of the detected objects are combined, along with the feature vector representation of the entire image to create a global representation of the image. We test three different object detection models: MASK-RCNN, YOLOv5, and YOLOv7 in order to find the best combination. 

In the second method, we only utilize the object encoder to obtain a feature vector representation of the image. We test various object encoders, including different versions of ResNet, DenseNet, EfficientNet, and CLIP. While object detection models have higher accuracy in detecting objects, they may not necessarily provide a more accurate feature vector representation of the entire image. By relying solely on the object encoder, we aim to capture a global representation that considers both the objects and their surroundings equally. The choice of an appropriate object encoder is crucial, as it is responsible for the feature vector used in our predictions. For both methods we calculate similarity using Cosine Similarity and use the scores to predict OOC/NOOC labels.

\section{Experiments}
\label{sec:experiments}
\subsection{Qualitative Analysis}
Defining image similarity is a difficult task, even for humans. One might consider a high-level comparison between images, such as the overall category or context of an image. For example, an image of an apple and a pear can be seen as similar because they are both fruits. On the other hand, taking a low-level comparison considering the colors, shapes and objects in the image, one might not find the images similar at all. In order to gather humanly annotated similarity scores, we conduct a survey using a small subset of the generated datasets, asking participants to rate the perceived similarity between the generated images for 24 caption pairs on a scale from 1-10, where 1 is the lowest degree of similarity and 10 the highest. We choose a sample of 24 image pairs (48 generated images) for each dataset with an even distribution of OCC/NOOC labels in the original caption pairs (12 of each). 

\begin{figure}
    \centering
    \includegraphics[width=\linewidth]{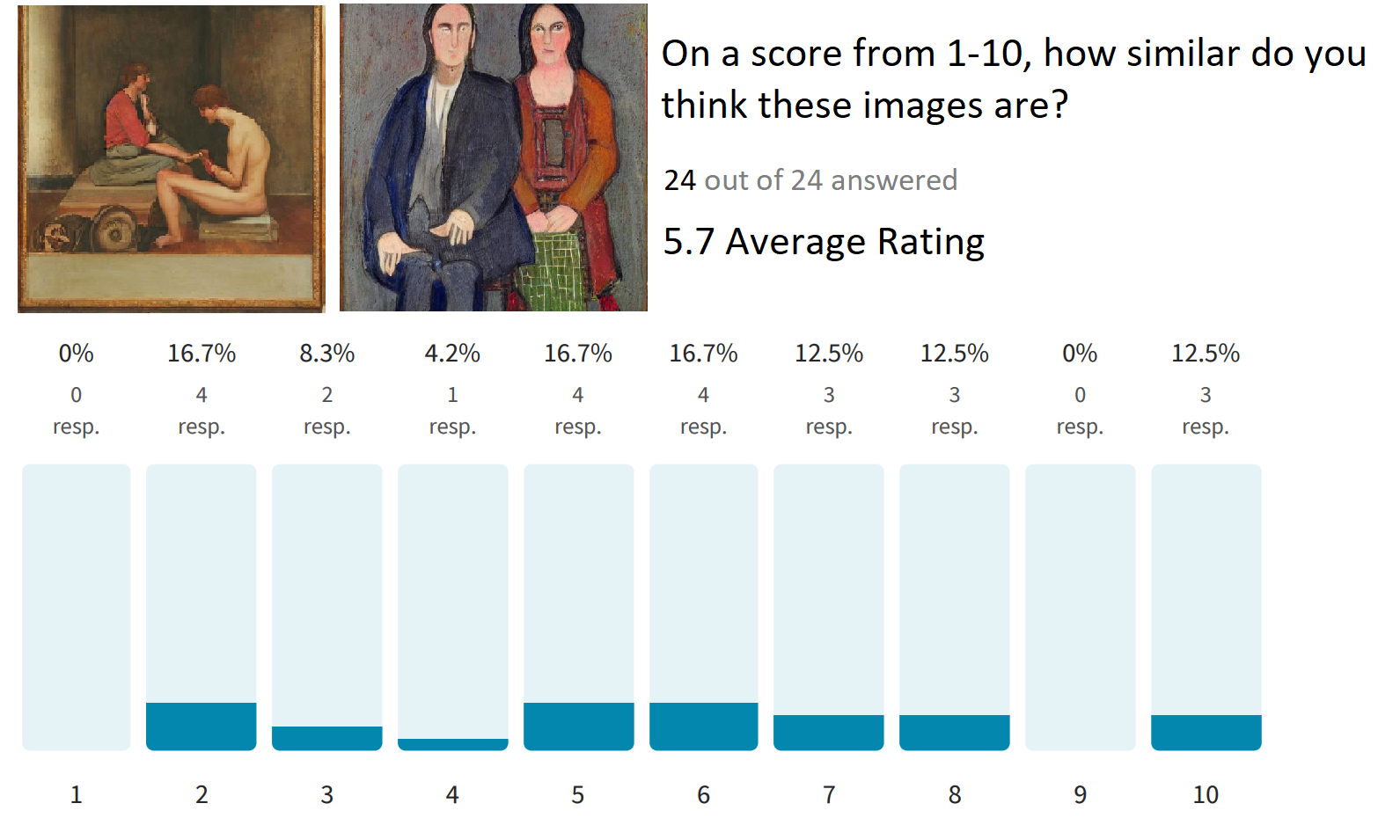}
    \caption[Similarity Score distributions in survey]{Distribution of similarity scores from survey. The distribution of ratings show a high variance in perceived similarity among the participants.}
    \label{fig:scoredist1}
\end{figure}

The average rating of the images indicate whether the participants find the image pairs more similar than dissimilar. On a scale of 1-10, 5.5 is the median rating. We can use this as a threshold as an indication as to whether participants believe the image are more similar than dissimilar or opposite. An average score equal to the threshold is defined as above. Our survey shows that rating the similarity between the images is a difficult task even for humans. The rating distributing shows a high variation in similarity scores for the same images. Figure \ref{fig:scoredist1} show an image pair where the variation of ratings is high, showing that the perceptual similarity of images highly varies from viewer to viewer. The caption pair used to generate the image pair in Figure \ref{fig:scoredist1} is NOOC. The average rating indicates that participants correctly identify this. This shows that the similarity in the image pair correlates to the similarity in the caption pair, and the text-to-image model effectively captures the semantic similarity.

Furthermore, we can convert the average ratings to corresponding OOC / NOOC labels in order to compare these to the gold labels from the COSMOS dataset for the caption pairs used to generate the images. Intuitively, an average score below the threshold indicates that the images are dissimilar, which in turn should indicate a presence of an OOC in the caption pairs used to generate the images. By defining scores below the threshold as OOC (1), and scores above or equal to the threshold as NOOC (0), we achieve the scores presented in table \ref{tab:surveyvmodelpredacc}. 

More importantly, the humanly annotated similarity scores allows us to evaluate whether our prediction model aligns with human perception, and allows us to analyze whether the prediction model accurately measures image similarity. By testing the variations of prediction models, we can see which model aligns with the predictions from the survey, presented in Table \ref{tab:surveyvmodelpredacc}.

\begin{table*}[t!]
\centering
\caption[Predictions with survey ratings versus model ratings]{Table shows the average ratings for the image pairs 1699-1676 in the survey converted to OOC/NOOC predictions, compared to the predictions from the models and the gold labels. 1 = OOC, 0 = NOOC.}
\label{tab:surveyvsCLIP}
\resizebox{\dimexpr\textwidth-1cm}{!}{%
\begin{tabular*}{\linewidth}{@{\extracolsep{\fill}} *{26}{l}}
\hline
\textbf{Image Pairs 16xx} & \textbf{99} & \textbf{98} & \textbf{97} & \textbf{96} & \textbf{95} & \textbf{94} & \textbf{93} & \textbf{92} & \textbf{91} & \textbf{90} & \textbf{89} & \textbf{88} & \textbf{87} & \textbf{86} & \textbf{85} & \textbf{84} & \textbf{83} & \textbf{82} & \textbf{81} & \textbf{80} & \textbf{79} & \textbf{78} & \textbf{77} & \textbf{76} \\ \hline
\textbf{CLIP SD} & \cellcolor[HTML]{34FF34}1, & \cellcolor[HTML]{34FF34}0, & \cellcolor[HTML]{34FF34}0, & \cellcolor[HTML]{FE0000}0, & \cellcolor[HTML]{34FF34}1, & \cellcolor[HTML]{34FF34}0, & \cellcolor[HTML]{34FF34}0, & \cellcolor[HTML]{FE0000}0, & \cellcolor[HTML]{34FF34}0, & \cellcolor[HTML]{34FF34}1, & \cellcolor[HTML]{34FF34}0, & \cellcolor[HTML]{FE0000}0, & \cellcolor[HTML]{FE0000}1, & \cellcolor[HTML]{FE0000}0, & \cellcolor[HTML]{34FF34}0, & \cellcolor[HTML]{34FF34}1, & \cellcolor[HTML]{34FF34}1, & \cellcolor[HTML]{34FF34}1, & \cellcolor[HTML]{FE0000}1, & \cellcolor[HTML]{34FF34}1, & \cellcolor[HTML]{FE0000}1, & \cellcolor[HTML]{FE0000}1, & \cellcolor[HTML]{34FF34}1, & \cellcolor[HTML]{FE0000}{\color[HTML]{333333} 1} \\ \hline
\textbf{SD Survey} & \cellcolor[HTML]{34FF34}1, & \cellcolor[HTML]{34FF34}0, & \cellcolor[HTML]{FE0000}1, & \cellcolor[HTML]{34FF34}1, & \cellcolor[HTML]{34FF34}1, & \cellcolor[HTML]{34FF34}0, & \cellcolor[HTML]{34FF34}0, & \cellcolor[HTML]{34FF34}1, & \cellcolor[HTML]{FE0000}1, & \cellcolor[HTML]{34FF34}1, & \cellcolor[HTML]{34FF34}0, & \cellcolor[HTML]{FE0000}0, & \cellcolor[HTML]{FE0000}1, & \cellcolor[HTML]{34FF34}1, & \cellcolor[HTML]{FE0000}{\color[HTML]{333333} 1,} & \cellcolor[HTML]{34FF34}1, & \cellcolor[HTML]{34FF34}1, & \cellcolor[HTML]{34FF34}1, & \cellcolor[HTML]{FE0000}1, & \cellcolor[HTML]{34FF34}1, & \cellcolor[HTML]{FE0000}1, & \cellcolor[HTML]{FE0000}1, & \cellcolor[HTML]{34FF34}1, & \cellcolor[HTML]{FE0000}{\color[HTML]{333333} 1} \\ \hline
\textbf{CLIP DALL-E 2} & \cellcolor[HTML]{34FF34}1, & \cellcolor[HTML]{34FF34}0, & \cellcolor[HTML]{FE0000}1, & \cellcolor[HTML]{34FF34}1, & \cellcolor[HTML]{34FF34}1, & \cellcolor[HTML]{34FF34}0, & \cellcolor[HTML]{FE0000}1, & \cellcolor[HTML]{FE0000}0, & \cellcolor[HTML]{34FF34}0, & \cellcolor[HTML]{34FF34}1, & \cellcolor[HTML]{34FF34}0, & \cellcolor[HTML]{FE0000}0, & \cellcolor[HTML]{34FF34}0, & \cellcolor[HTML]{FE0000}0, & \cellcolor[HTML]{34FF34}0, & \cellcolor[HTML]{34FF34}1, & \cellcolor[HTML]{34FF34}1, & \cellcolor[HTML]{34FF34}1, & \cellcolor[HTML]{FE0000}1, & \cellcolor[HTML]{34FF34}1, & \cellcolor[HTML]{34FF34}0, & \cellcolor[HTML]{FE0000}1, & \cellcolor[HTML]{FE0000}0, & \cellcolor[HTML]{FE0000}{\color[HTML]{333333} 1} \\ \hline
\textbf{DALL-E 2 Survey} & \cellcolor[HTML]{34FF34}1, & \cellcolor[HTML]{34FF34}0, & \cellcolor[HTML]{FE0000}1, & \cellcolor[HTML]{34FF34}1, & \cellcolor[HTML]{34FF34}1, & \cellcolor[HTML]{34FF34}0, & \cellcolor[HTML]{FE0000}1, & \cellcolor[HTML]{34FF34}1, & \cellcolor[HTML]{34FF34}0, & \cellcolor[HTML]{34FF34}1, & \cellcolor[HTML]{FE0000}1, & \cellcolor[HTML]{34FF34}1, & \cellcolor[HTML]{FE0000}1, & \cellcolor[HTML]{FE0000}0, & \cellcolor[HTML]{34FF34}0, & \cellcolor[HTML]{34FF34}1, & \cellcolor[HTML]{34FF34}1, & \cellcolor[HTML]{34FF34}1, & \cellcolor[HTML]{FE0000}1, & \cellcolor[HTML]{34FF34}1, & \cellcolor[HTML]{FE0000}1, & \cellcolor[HTML]{FE0000}1, & \cellcolor[HTML]{34FF34}1, & \cellcolor[HTML]{FE0000}{\color[HTML]{333333} 1} \\ \hline
\textbf{Gold labels} & \cellcolor[HTML]{34FF34}1, & \cellcolor[HTML]{34FF34}0, & \cellcolor[HTML]{34FF34}0, & \cellcolor[HTML]{34FF34}1, & \cellcolor[HTML]{34FF34}1, & \cellcolor[HTML]{34FF34}0, & \cellcolor[HTML]{34FF34}0, & \cellcolor[HTML]{34FF34}1, & \cellcolor[HTML]{34FF34}0, & \cellcolor[HTML]{34FF34}1, & \cellcolor[HTML]{34FF34}0, & \cellcolor[HTML]{34FF34}1, & \cellcolor[HTML]{34FF34}0, & \cellcolor[HTML]{34FF34}1, & \cellcolor[HTML]{34FF34}0, & \cellcolor[HTML]{34FF34}1, & \cellcolor[HTML]{34FF34}1, & \cellcolor[HTML]{34FF34}1, & \cellcolor[HTML]{34FF34}0, & \cellcolor[HTML]{34FF34}1, & \cellcolor[HTML]{34FF34}0, & \cellcolor[HTML]{34FF34}0, & \cellcolor[HTML]{34FF34}1, & \cellcolor[HTML]{34FF34}0 \\ \hline
\end{tabular*}%
}
\end{table*}

\begin{table}[h!]
\centering
\caption[Survey vs model prediction accuracy.]{As shown in Table \ref{tab:surveyvsCLIP}, the survey and model similarity ratings correlate. This table shows the prediction accuracy and precision the data in Table \ref{tab:surveyvsCLIP} converts to, and also the accuracy score of the other encoders.}
\label{tab:surveyvmodelpredacc}
\begin{tabular}{l|cc|cc|}
\cline{2-5}
 & \multicolumn{2}{c|}{\textbf{SD}} & \multicolumn{2}{c|}{\textbf{DALL-E 2}} \\ \cline{2-5} 
 & \multicolumn{1}{l|}{\textbf{Accuracy}} & \multicolumn{1}{l|}{\textbf{Precision}} & \multicolumn{1}{l|}{\textbf{Accuracy}} & \multicolumn{1}{l|}{\textbf{Precision}} \\ \hline
\multicolumn{1}{|l|}{\textbf{Survey}} & \multicolumn{1}{c|}{{\color[HTML]{3166FF} \textbf{0.63}}} & {\color[HTML]{3166FF} \textbf{0.58}} & \multicolumn{1}{c|}{{\color[HTML]{3166FF} \textbf{0.63}}} & {\color[HTML]{3166FF} \textbf{0.58}} \\ \hline
\multicolumn{1}{|l|}{\textbf{CLIP}} & \multicolumn{1}{c|}{{\color[HTML]{3166FF} \textbf{0.63}}} & {\color[HTML]{3166FF} \textbf{0.62}} & \multicolumn{1}{c|}{{\color[HTML]{3166FF} \textbf{0.63}}} & {\color[HTML]{3166FF} \textbf{0.62}} \\ \hline
\multicolumn{1}{|l|}{\textbf{ResNet18}} & \multicolumn{1}{c|}{0.50} & 0.50 & \multicolumn{1}{c|}{0.45} & 0.46 \\ \hline
\multicolumn{1}{|l|}{\textbf{ResNet50}} & \multicolumn{1}{c|}{0.50} & 0.50 & \multicolumn{1}{c|}{0.45} & 0.45 \\ \hline
\multicolumn{1}{|l|}{\textbf{ResNext50}} & \multicolumn{1}{c|}{0.54} & 0.53 & \multicolumn{1}{c|}{0.50} & 0.50 \\ \hline
\multicolumn{1}{|l|}{\textbf{DenseNet121}} & \multicolumn{1}{c|}{0.41} & 0.41 & \multicolumn{1}{c|}{0.45} & 0.45 \\ \hline
\multicolumn{1}{|l|}{\textbf{DenseNet169}} & \multicolumn{1}{c|}{0.45} & 0.47 & \multicolumn{1}{c|}{0.58} & 0.55 \\ \hline
\multicolumn{1}{|l|}{\textbf{EfficientNet}} & \multicolumn{1}{c|}{0.54} & 0.53 & \multicolumn{1}{c|}{0.58} & 0.56 \\ \hline
\end{tabular}
\end{table}

\subsection{Quantitative Analysis}
\begin{table}[t]
\centering
\caption[Difference in Object Detection Model performance.]{Difference in Object Detection Model performance. YOLOv7 outperforms the other models, but do not outperform ResNet50 on its own. The predictions are based on Gen vs Gen similarity on the Stable Diffusion dataset.}
\resizebox{\linewidth}{!}{%
\begin{tabular}{c|cccc|}
\cline{2-5}
 & \multicolumn{4}{c|}{\textbf{Object detection + ResNet50}} \\ \cline{2-5} 
 & \multicolumn{1}{c|}{\textbf{ResNet50}} & \multicolumn{1}{c|}{\textbf{+YOLOv7}} & \multicolumn{1}{c|}{\textbf{+MASK-RCNN}} & \textbf{+ YOLOv5} \\ \hline
\rowcolor[HTML]{C0C0C0} 
\multicolumn{1}{|c|}{\cellcolor[HTML]{C0C0C0}\textbf{Accuracy:}} & \multicolumn{1}{c|}{\cellcolor[HTML]{C0C0C0}{\color[HTML]{3166FF} \textbf{0.60}}} & \multicolumn{1}{c|}{\cellcolor[HTML]{C0C0C0}0.54} & \multicolumn{1}{c|}{\cellcolor[HTML]{C0C0C0}0.53} & 0.50 \\ \hline
\multicolumn{1}{|c|}{\textbf{Precision:}} & \multicolumn{1}{c|}{{\color[HTML]{3166FF} \textbf{0.60}}} & \multicolumn{1}{c|}{0.55} & \multicolumn{1}{c|}{0.53} & 0.50 \\ \hline
\rowcolor[HTML]{C0C0C0} 
\multicolumn{1}{|c|}{\cellcolor[HTML]{C0C0C0}\textbf{Recall:}} & \multicolumn{1}{c|}{\cellcolor[HTML]{C0C0C0}{\color[HTML]{3166FF} \textbf{0.60}}} & \multicolumn{1}{c|}{\cellcolor[HTML]{C0C0C0}0.53} & \multicolumn{1}{c|}{\cellcolor[HTML]{C0C0C0}0.53} & 0.50 \\ \hline
\multicolumn{1}{|c|}{\textbf{F1}} & \multicolumn{1}{c|}{{\color[HTML]{3166FF} \textbf{0.60}}} & \multicolumn{1}{c|}{0.54} & \multicolumn{1}{c|}{0.53} & 0.50 \\ \hline
\end{tabular}%
}
\label{tab:yolov7accuracy}
\end{table}

Quantitative analysis is performed using the automated prediction model. We test a total of 8 object encoders paired with 3 object detection models. While MASK-RCNN and YOLOv7 has better detection accuracy than YOLOv5, it does not improve the performance of the model significantly. Table \ref{tab:yolov7accuracy} shows the accuracy score of the model when utilizing YOLOv7 over other object detection models with the ResNet50 encoder. YOLOv7 produces a slightly better detection accuracy when paired with ResNet50 than YOlOv5 and MASK-RCNN. However, the slight increase in accuracy comes with a huge increase in runtime when utilizing normal GPUs. While the other variations use around 30 minutes on prediction on the entire dataset, YOLOv7 uses around 1hour and 30 minutes. MASK-RCNN, despite boosting better detection accuracy than YOLOv5, actually performs worse than YOLOv5 paired with all object encoders expect for EfficentNet on the DALL-E 2 dataset, where it returns a 1\% better accuracy. However, on the Stable Diffusion dataset, utilizing MASK-RCNN is superior to YOLOv5 and provides a 5-7\% boost in accuracy in general. The runtime is also similar on a normal Colab GPU. The best performing version of our model utilizing an object detection model is MASK-RCNN combined with an EfficientNet-B5 object encoder, yielding a 0.57\% detection accuracy. 

Despite this, none of the versions where we utilize object detection models outperform the versions where we only utilize object encoders. Paired with our best performing model, YOLOv5 decreases the accuracy score of the CLIP model by 16\% on the Stable Diffusion dataset. We see a general 10\% accuracy decrease when utilizing object detection models, versus only utilizing object encoders. Therefore, it is a clear advantage of utilizing only object encoders for this task, both in terms of accuracy and runtime. Our study shows that several object encoders are able to accurately capture the perceptual similarity between images without the need for additional detection methods. The scores are presented in Table \ref{tab:overallaccuracy} and \ref{tab:bestscores}. The performance difference between DALL-E 2 and Stable Diffusion is negligible, and the accuracy scores are mostly closely matched across various encoders, with slight variations. 

Setting the right conditional rule for predictions is a difficult task. We calculate similarity of both generated images towards the original image, achieving two similarity scores, sim1 and sim2. We set a threshold for OOC/NOOC prediction of 0.50. Intuitively we can rule that if both images fall below the similarity threshold, we predict OOC, while if both images are above, we predict NOOC. However, this method presents a challenge as the generated images may not be comparable to the original images despite being NOOC. The modified captions used as prompts for the text-to-image generative models may differ from the original captions, so that both generated images are different towards the original image, despite both the original captions conveying the same semantic meaning. Hence, we might assume that if the generated images are similar, i.e. if both similarity scores either exceed or fall below the threshold, they are deemed as NOOC. We find that a combination of the two if/else statements mentioned above yields the most effective results. We utilize an if/else statement to capture if both sim1 and sim2 are below the threshold, predicting OOC. Furthermore, we can utilize an elif statement to rule that if sim1 is below the threshold and sim2 is above the threshold, or Opposite, we also predict OOC. If none of the previous conditions are met, this means that both sim1 and sim2 are on the same side of the threshold, having an equal similarity towards the original. If so, a NOOC label is predicted.

\begin{table}[]
\centering
\caption[CLIP model prediction scores]{Predictions using only CLIP model for feature vector representations. We use a median threshold for predictions when comparing generated images. The CLIP model gains the best overall performance on both datasets. The best scores are outlined in blue.}
\begin{tabular}{c|cc|cc|}
\cline{2-5}
 & \multicolumn{2}{c|}{\textbf{CLIP ViT-L-14}} & \multicolumn{2}{c|}{\textbf{CLIP ViT-B-32}} \\ \cline{2-5} 
 & \multicolumn{1}{c|}{\textbf{SD}} & \textbf{DALL-E 2} & \multicolumn{1}{c|}{\textbf{SD}} & \textbf{DALL-E 2} \\ \hline
\rowcolor[HTML]{C0C0C0} 
\multicolumn{1}{|c|}{\cellcolor[HTML]{C0C0C0}\textbf{Accuracy}} & \multicolumn{1}{c|}{\cellcolor[HTML]{C0C0C0}0.664} & {\color[HTML]{3166FF} \textbf{0.682}} & \multicolumn{1}{c|}{\cellcolor[HTML]{C0C0C0}0.638} & 0.668 \\ \hline
\multicolumn{1}{|c|}{\textbf{Precision}} & \multicolumn{1}{c|}{0.678} & 0.68 & \multicolumn{1}{c|}{0.656} & {\color[HTML]{3166FF} \textbf{0.689}} \\ \hline
\rowcolor[HTML]{C0C0C0} 
\multicolumn{1}{|c|}{\cellcolor[HTML]{C0C0C0}\textbf{Recall}} & \multicolumn{1}{c|}{\cellcolor[HTML]{C0C0C0}0.626} & {\color[HTML]{3166FF} \textbf{0.687}} & \multicolumn{1}{c|}{\cellcolor[HTML]{C0C0C0}0.582} & 0.614 \\ \hline
\multicolumn{1}{|c|}{\textbf{F1}} & \multicolumn{1}{c|}{0.651} & {\color[HTML]{3166FF} \textbf{0.683}} & \multicolumn{1}{c|}{0.617} & 0.649 \\ \hline   \end{tabular}
\label{tab:bestscores}
\end{table}

\begin{table}[tb]
\centering
\caption[Accuracy Scores for each model combination]{Accuracy scores for each combination of object encoder or object encoder + object detection model.}
\resizebox{\linewidth}{!}{%
\begin{tabular}{c|ccc|ccc|}
\cline{2-7}
 & \multicolumn{3}{c|}{\textbf{DALL-E 2}} & \multicolumn{3}{c|}{\textbf{SD}} \\ \cline{2-7} 
 & \multicolumn{1}{c|}{\textbf{Encoder}} & \multicolumn{1}{c|}{\textbf{M-RCNN}} & \textbf{YOLOv5} & \multicolumn{1}{c|}{\textbf{Encoder}} & \multicolumn{1}{c|}{\textbf{M-RCNN}} & \textbf{YOLOv5} \\ \hline
 \rowcolor[HTML]{C0C0C0} 
\multicolumn{1}{|c|}{\textbf{EfficientNet-B5}} & \multicolumn{1}{c|}{{\color[HTML]{3166FF} \textbf{0.56}}} & \multicolumn{1}{c|}{{\color[HTML]{3531FF} \textbf{0.51}}} & 0.50 & \multicolumn{1}{c|}{{\color[HTML]{3166FF} \textbf{0.60}}} & \multicolumn{1}{c|}{{\color[HTML]{3166FF} \textbf{0.58}}} & 0.50 \\ \hline
\multicolumn{1}{|c|}{\textbf{ResNet18}} & \multicolumn{1}{c|}{{\color[HTML]{3166FF} \textbf{0.51}}} & \multicolumn{1}{c|}{0.46} & 0.50 & \multicolumn{1}{c|}{{\color[HTML]{3166FF} \textbf{0.60}}} & \multicolumn{1}{c|}{{\color[HTML]{212121} 0.55}} & 0.50 \\ \hline
\rowcolor[HTML]{C0C0C0} 
\multicolumn{1}{|c|}{\textbf{ResNet50}} & \multicolumn{1}{c|}{{\color[HTML]{3166FF} \textbf{0.52}}} & \multicolumn{1}{c|}{0.45} & 0.48 & \multicolumn{1}{c|}{{\color[HTML]{3166FF} \textbf{0.60}}} & \multicolumn{1}{c|}{{\color[HTML]{212121} 0.52}} & 0.49 \\ \hline
\multicolumn{1}{|c|}{\textbf{ResNext50}} & \multicolumn{1}{c|}{{\color[HTML]{3166FF} \textbf{0.51}}} & \multicolumn{1}{c|}{0.47} & 0.48 & \multicolumn{1}{c|}{{\color[HTML]{3166FF} \textbf{0.59}}} & \multicolumn{1}{c|}{{\color[HTML]{212121} 0.55}} & 0.49 \\ \hline
\rowcolor[HTML]{C0C0C0} 
\multicolumn{1}{|c|}{\textbf{DenseNet121}} & \multicolumn{1}{c|}{{\color[HTML]{3166FF} \textbf{0.51}}} & \multicolumn{1}{c|}{0.48} & 0.49 & \multicolumn{1}{c|}{{\color[HTML]{3166FF} \textbf{0.58}}} & \multicolumn{1}{c|}{{\color[HTML]{212121} 0.56}} & 0.50 \\ \hline
\multicolumn{1}{|c|}{\textbf{DenseNet169}} & \multicolumn{1}{c|}{{\color[HTML]{3166FF} \textbf{0.54}}} & \multicolumn{1}{c|}{0.48} & 0.50 & \multicolumn{1}{c|}{{\color[HTML]{3166FF} \textbf{0.59}}} & \multicolumn{1}{c|}{{\color[HTML]{212121} 0.56}} & 0.50 \\ \hline
\end{tabular}%
}
\label{tab:overallaccuracy}
\end{table}

\begin{figure}
    \centering
    \includegraphics[scale=0.25]{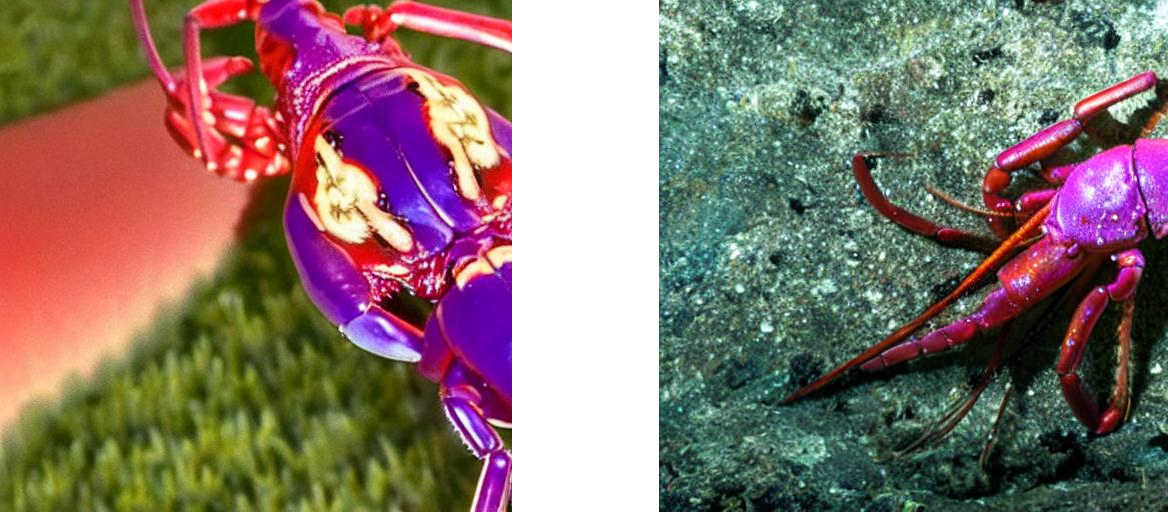}
    \caption{Example of false positive: Generated image pair based on caption pair:
    1:"A photograph shows a purple lobster caught in Maine," (left). 
    2: "'One-in-a-million' purple lobster fools the internet since it is not genuine" (right).}
    \label{fig:purplelobster}
\end{figure}

\section{Discussion}

\begin{figure}[ht]
  \centering
  \includegraphics[width=\linewidth]{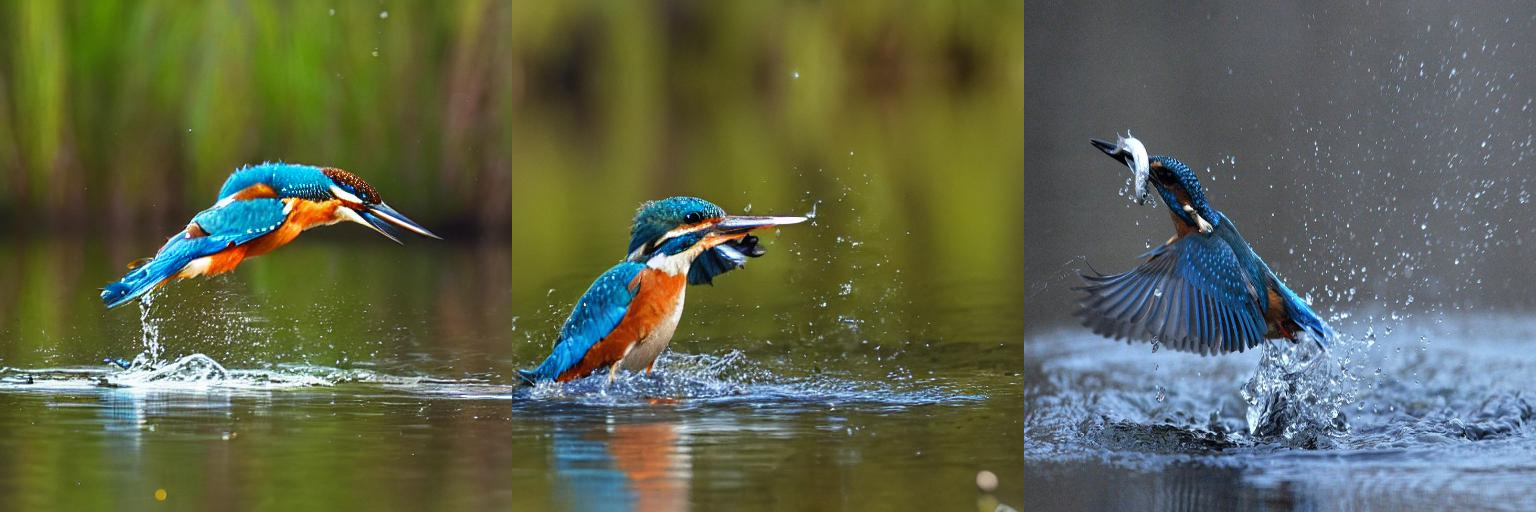}
  \caption[High similarity between generated image pairs and original image.]{Generated image pair with a high similarity resemblance towards the original image suggest the models ability to create highly realistic and semantically aligned images from descriptive captions. The original image is to the right.}
    \label{fig:birdfish}
\end{figure}

\begin{figure}[ht]
    \centering
    \includegraphics[scale=0.25]{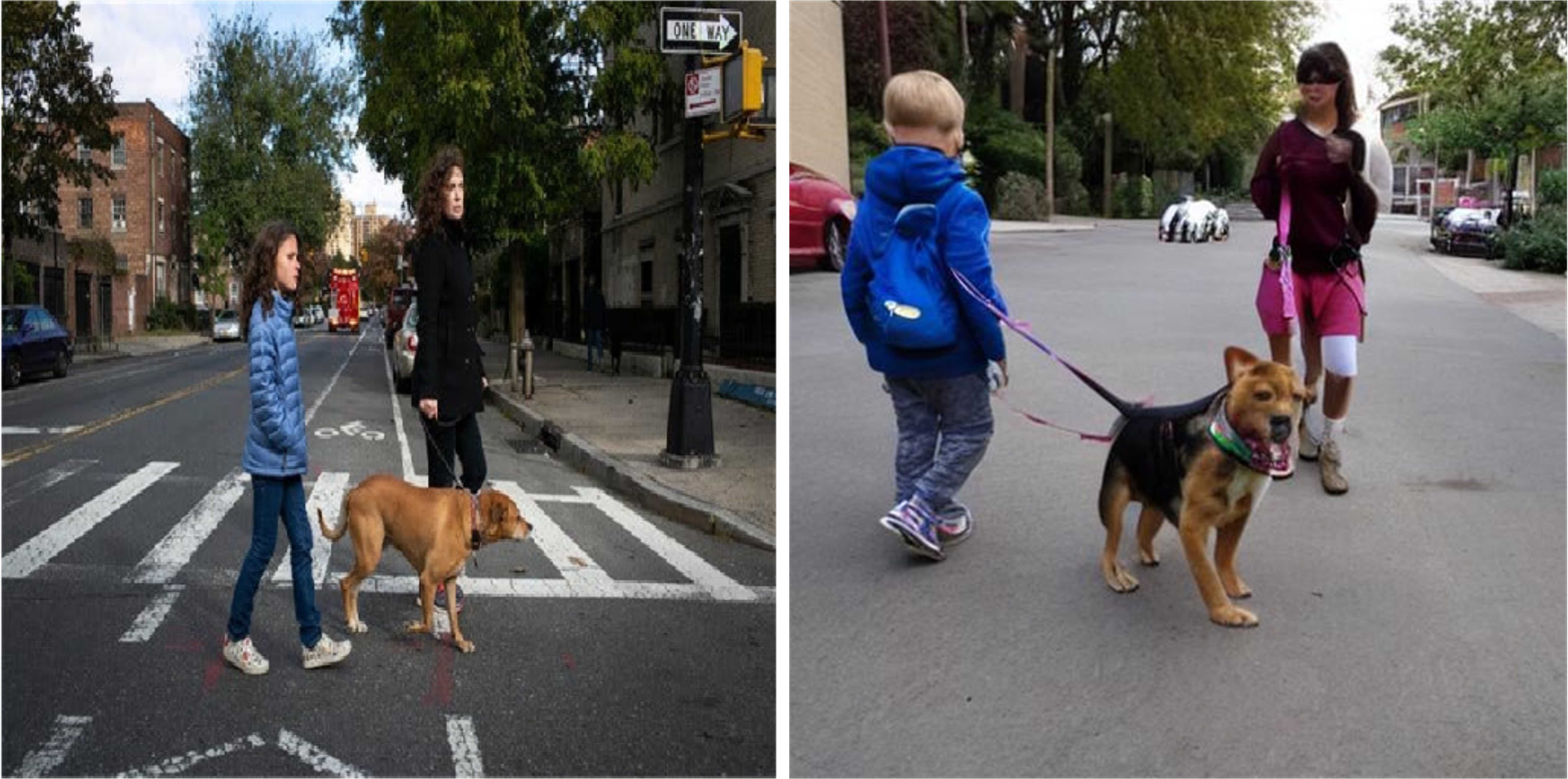}
    \caption[Similarity of original vs generated image 1676.]{Original image 1676 vs generated caption 2: ´I'm primarily the dog walker, but usually the kids come with me´. The original image is to the left.}
    \label{fig:Original1676}
\end{figure}

The proposed method is subject to a range of factors, not only the actual performance of the generative models for the task, but also the quality of the captions used to generate images, how well the feature vector representations captures the content of the image, and the way predictions are carried out and what similarity measures are employed. Consequently there are numerous opportunities for optimization, and finding the most optimal approach is not necessarily straight forward. In order to find an effective combination, we test a variety of object encoders and object detection models. The analysis proves that utilizing image generation models for this task provides both strengths and limitations.

Figure \ref{fig:birdfish} demonstrates that the models generate highly realistic images that closely resembles the original image, allowing us to make correct predictions. Figure \ref{fig:Original1676} also shows that the models are able to generate images with high semantic alignment to the input caption. This shows that the image generation models can effectively capture the semantic similarity or dissimilarity in news captions. 


Using the survey scores with humanly annotated similarity ratings, we are able to compare how the models align with human perception, and as such, which one of our model versions most objectively capture perceptual similarity. Table \ref{tab:surveyvsCLIP} shows how CLIP strongly correlates to human perception on the image pairs in the subset used in the survey. Table \ref{tab:surveyvmodelpredacc} shows how the predictions converted to accuracy and precision scores, and demonstrates each models effectiveness at capturing accurate similarity scores and alignment with human perception. This serves as an indicator that our version utilizing CLIP accurately measures the efficiency of Stable Diffusion and DALL-E 2 for detecting mismatch in image caption pairs. We find that CLIP presents a high correlation with human annotations, suggesting CLIP aligns with human perception, even for difficult tasks such as the subset utilized in the survey. Therefore we are confident that the reported scores from our model version utilizing CLIP provides an accurate and reliable evaluation of the performance of both image generation models for the task of cheapfake detection.


The qualitative analysis demonstrates that the models are not able to capture contradictions in the caption pairs. This also assessed by Marcus \textit{et. al} for DALL-E 2 \cite{marcus2022preliminary}. Akgul \textit{et al.} and Tran \textit{et al.} demonstrate the increase in accuracy achieved when detecting contradictions in the caption pairs \cite{akgul2021cosmos}\cite{tran2022textual}. The COSMOS baseline approach defines that captions referring to different objects in the image are considered NOOC. NOOC captions often contain references to different objects or describe the same object differently. Our method does not incorporate any rules prior to image generation. Instead, we directly feed the captions to the image generation models. Combining previously proposed methods for cheapfake detection could therefore effectively increase the performance of our approach.


\section{Conclusion}

Our paper conducts a comprehensive analysis of DALL-E 2 and Stable Diffusion in generating news-related images. The rapid progress in AI generative models presents opportunities for further research. For example, Midjourney has shown exceptional image generation capabilities \cite{mitchell_2023} \cite{Midjourney}. Google's Imagen achieves state-of-the-art performance on the COCO dataset and is preferred by human evaluators compared to other models \cite{saharia2022photorealistic}. We recommend exploring these new models when they become publicly available or accessible through APIs.

Additionally, there are optimization steps to enhance the model's accuracy in future research. The text pre-processing step may result in decreased contextual information, particularly for non-descriptive captions where entity labels replace most of the words, leading to a loss of context. Certain NER tags like LOC and GPE are difficult to interpret, as well as ambiguous words like DATE and CARDINAL, posing challenges for DALL-E 2 and Stable Diffusion. NER tagging for dates and numbers may not be necessary as they do not introduce harmful context triggering safety filters or harmful content in the datasets. However, it's important to acknowledge that navigating the safety filters of DALL-E 2 and Stable Diffusion is challenging. Optimizing the text-processing step is crucial for achieving higher performance, but it requires careful consideration of the safety filters in both models.
\begin{acks}
This research was funded by NORDIS, European Horizon 2020 grant number 825469.
\end{acks}

\bibliographystyle{ACM-Reference-Format}
\bibliography{references}


\end{document}